# Comparison of Different Methods for Time Sequence Prediction in Autonomous Vehicles

Teng Liu
Vehicle Intelligence Pioneers Inc.
Qingdao Academy of Intelligent Industries
Qingdao Shandong 266109, China
tengliu17@ gmail.com

Bin Tian
Vehicle Intelligence Pioneers Inc.
Qingdao Academy of Intelligent Industries
Qingdao Shandong 266109, China
bin.tian@ia.ac.cn bin.tian@ia.ac.cn

Yunfeng Ai
Vehicle Intelligence Pioneers Inc.
Qingdao Academy of Intelligent Industries
Qingdao Shandong 266109, China
aiyunfeng@ucas.ac.cn

Long Chen
Vehicle Intelligence Pioneers Inc.
School of Data and Computer Science
Sun Yat-sen University
Guangzhou, China
chenl46@mail.sysu.edu.cn

Fei Liu
Xilinx Technology Beijing Limited
Beijing, 100083 China
fledaliu@xilinx.com

Dongpu Cao
Department of Mechanical and Mechatronics Engineering
University of Waterloo
Ontario N2L3G1, Canada
dongpu@uwaterloo.ca

*Abstract*—As a combination of various kinds of technologies, autonomous vehicles could complete a series of driving tasks by itself, such as perception, decision-making, planning and control. Since there is no human driver to handle the emergency situation, future transportation information is significant for automated vehicles. This paper proposes different methods to forecast the time series for autonomous vehicles, which are the nearest neighborhood (NN), fuzzy coding (FC) and long short term memory (LSTM). First, the formulation and operational process for these three approaches are introduced. Then, the vehicle velocity is regarded as a case study and the real-world dataset is utilized to predict future information via these techniques. Finally, the performance, merits and drawbacks of the presented methods are analyzed and discussed.

## I. Introduction

Owing to the explosive development of information transmission and intelligent equipment, the smart city is gradually realized in our daily life [1]. Autonomous vehicles are treated as an important part of the Intelligent Transportation System (ITS) due to their high mobility and convenience. Four modules are contributed to enable automated vehicles to achieve various kinds of driving tasks [2], i.e., perception, decision-making, planning and control. Furthermore, future information is significant for autonomous vehicles to improve reliability and handle emergency [3].

Many physical quantities in vehicles can be regarded as time series, and thus serval approaches have been presented to forecast the future information. For example, in Ref. [4], the authors developed a steady-state Kalman filter to predict and estimate the lateral force and yaw moment. The proposed model could improve the accuracy of path prediction when external disturbances existed. Xie *et al.* combined maneuver- and physics-based methods to forecast the future trajectory for automated vehicles [5]. This integrated model is validated based on the naturalistic driving data with higher accuracy and longer horizon. To estimate the time-to-collision (TTC) precisely, Kim *et al.* presented a threat prediction algorithm in [6]. The surrounding vehicles trajectories and local path candidates are modeled and predicted to decide the vehicle's driving strategy. In Ref. [7], the authors applied model predictive control (MPC) to consider and predict probable risky behaviors of the surrounding vehicles. By doing this, the automated vehicles can be operated in a safe driving envelope. Furthermore, Wiest et al. employed Bernoulli-Gaussian mixture model to forecast the behavioral maneuvers at intersections [8]. Then, this model is extended for multiple intersections through online learning.

Besides the conventional techniques, the deep learning (DL) methods became prevalent in time series prediction recently. For example, a DL-based traffic flow prediction approach is presented in [9], wherein a stacked autoencoder (SAE) model is used to learn the generic features. Experiments prove the proposed prediction model is satisfying for real-world applications. Ref. [10] discusses a DL framework for human motion prediction. To represent the features of human motions, the authors employ the encoding-decoding network to forecast the future 3D poses. For video prediction in Atari games, Oh *et al.* proposed two deep neural network architectures to predict the future image-frames [11]. Tests demonstrated the generated visually-realistic frames were suitable for about 100-step prediction. Furthermore, DL technique is also applied for naturalistic speech prediction and generation. The authors in [12] utilized a sliding window predictor to learn the mapping from phoneme label sequences to mouth movements. Then, the end-to-end speech animation is created using a deep neural network.

To achieve accurate and effective time series prediction for autonomous vehicles, this paper presents three different approaches to realize this aim. They are the nearest neighborhood (NN), fuzzy coding (FC) and long short term memory (LSTM). These methods are feasible for most of the variables in a vehicle, such as vehicle speed, acceleration, power demand, engine speed and torque, human driver's intention and behaviors. Without loss of generality, this work takes the vehicle velocity as a case study and applies these three approaches to forecast the future vehicle speed based on the real-world driving data. The mathematics and operational process are first discussed, and then the performance of the presented methods are evaluated. The merits and drawbacks of these methods are finally illustrated.

The construction of the rest of the paper is organized as follows: the transition probability-based NN and FC are given in Section II. The special LSTM network in DL framework is formulated in Section III. Experiment results are analyzed in Section IV, and Section V concludes the paper.

## II. Nearest Neighborhood and Fuzzy Coding Methods

This section explains the transition probability-based NN and FC methods. The maximum likelihood estimator (MLE) is first introduced to calculate the transition probability. Then, the mathematics of the conventional NN and FC techniques are displayed for one step and multi-step prediction.

### A. Nearest Neighborhood

To predict the time sequence in automated vehicles, the variable interval is divided into finite discrete states, which are described as $X=\{x_i | i=1, …, M\}$ and subject to $\{ x_i < x_{i+1} | \forall i \}$. For an arbitrary continuous time sequence, it can be mapped to these discrete states via nearest neighborhood measurement as follow

$$y = x_j, j = \arg\min_i |y - x_i| \quad (1)$$

where $y$ denotes a special time sequence at arbitrary time point. After the mapping is established, a Markov chain (MC) is usually used to mimic the discrete states. Then, the transition probability of these states are computed via maximum likelihood estimator (MLE) as

$$p_{ij} = \frac{N_{ij}}{N_{io}}, \ i,j \in \{1,...,M\} \quad (2)$$

where $p_{ij}$ represents the transition probability from $x_i$ to $x_j$. $N_{ij}$ and $N_{io}$ are the transition counts from $x_i$ to $x_j$ and $x_i$ to all the objectives, respectively. Thus, these two numbers are satisfied by

$$N_{io} = \sum_{j=1}^{M} N_{ij}, \ i,j \in \{1,...,M\} \quad (3)$$

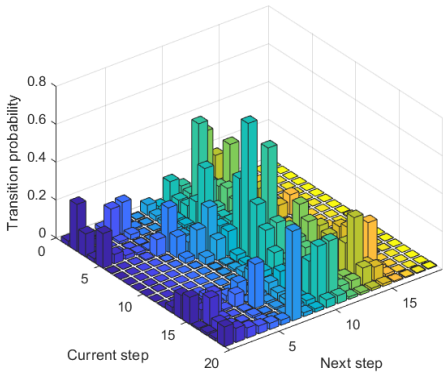

Fig. 1. Example of computed transition probability for time sequence.

The obtained example of transition probability is shown in Fig. 1. Based on the obtained transition probability, the future one-step and multi-step variables can be predicted in two forms. The first one is using the maximum probability to compute the future variable as

$$x_{t+1} = x_k, \ if \ x_t = x_i, k = \arg\max_k (p_{ik}) \quad (4)$$

The second idea is exploiting the expectation of the next state to depict the predicted value as

$$x_{t+1} = \sum_{k=1}^{M} p_{ik} \cdot x_k, \ if \ x_t = x_i \quad (5)$$

It is obvious that the second expression is more accurate that the first one because it considers all the possible transitions. In this work, the second formulation is utilized to represent the NN-based prediction. Finally, the multi-step future variables are computed by extending (5) as follow

$$x_{t+n} = \sum_{k=1}^{M} (p_{ik})^n \cdot x_k, \ if \ x_t = x_i \quad (6)$$

### B. Fuzzy Coding

In FC method, the discrete signal states $x_i$ are replaced by fuzzy sets $A_i$, $i=1, …, M$. Each set $A_i$ is a pair of $X$ and $\mu_i$, wherein $\mu_i$ is called as Lebesgue measurable membership function and defined as

$$\mu_i : X \rightarrow [0,1], \ s.t. \ \forall y, \exists i, \mu_i(y) > 0 \quad (7)$$

Hence, $\mu_i$ can be interpreted as the degree of $y \in A_i$. Since this membership maps $X$ to $[0, 1]$, a transformation needs to be executed from arbitrary variable $y$ to a possibility vector as

$$\mu^T(y) = [\mu_1(y), \mu_2(y), \cdots, \mu_M(y)] \quad (8)$$

As the $\mu_i(y)$ represents the degree of membership, the sum of the elements in $\mu^T(y)$ may not equal to 1. It is because that each variable $y$ could be associated with more than one fuzzy set $A_i$.

Furthermore, another transformation is required to map the possibility vector to a probability vector as follow

$$\theta_i(y) = \mu_i(y) / \sum_{j=1}^{M} \mu_j(y), \ i = 1,...M \quad (9)$$

Finally, the FC-based next-step variable is computed by considering the probability vector $\theta_i(y)$, transition probability $p_{ij}$ and membership function $\mu_i(y)$ together [13]

$$y_{t+1} = \frac{\sum_{i=1}^{M} \theta_i(y) \sum_{j=1}^{M} p_{ij} \int_X y\mu_j(y) dy}{\sum_{i=1}^{M} O_i(y) \sum_{j=1}^{M} p_{ij} \int_X \mu_j(y) dy} \quad (10)$$

where $\theta_i(y)$ and $p_{ij}$ are probability and thus the sum of them are equal to 1. In this work, the membership function $\mu_i(y)$ are specialized as the Gaussian function with variance $\sigma=1$

$$\ln \mu_i = -\frac{(x - 2.5i + 1.25)^2}{2 \cdot \sigma^2}, \quad i = 1,...M \quad (11)$$

## III. LONG SHORT TERM MEMORY NETWORK

In this section, the LSTM-based time series prediction framework is constructed. First, the artificial neural network (ANN) and recurrent neural network (RNN) are introduced. Then, the mathematics of the LSTM network is formulated, wherein the input, forget and output cells are described in detail. Finally, the real-world driving dataset called Next Generation Simulation (NGSIM) dataset used for time sequence prediction is depicted.

### A. Recurrent Neural Network

Deep learning (DL) is a hot research topic in recent years, which is inspired by learning features from collected data representations [14]. DL usually contains several levels and each level aims to transform the coarse data into more composite and abstract representation. Generally, the artificial neural network (ANN) is used to mimic the DL models and it can be indicated as function approximation between the input and output. This approximation is often achieved by three common layers, which are input, hidden and output layers [15].

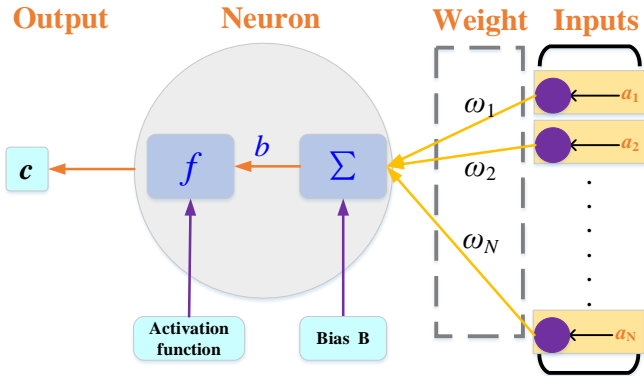

Fig. 2. The architecture of the neuron unit in ANN.

The neuron is a sole computation unit of ANN, see Fig. 2 as an illustration. The input and weight vectors are denoted as $a=[a_1, a_2, ...a_N]$ and $\omega=[\omega_1, \omega_2,...\omega_N]$, and then the neuron input $b$ is depicted as

$$b = \sum_{i=1}^{N} a_i \cdot \omega_i + B \quad (12)$$

where $B$ is a bias. The activation function is acted on the neuron input to generate the restricted and squashed output value

$$c = f(b) \quad (13)$$

wherein the activation function $f$ can be Sigmoid, Tanh and Rectified Linear Unit (Relu) function.

In many cases, the order of inputs will influence the outputs, such as the natural language data, speech and music sequence data. Recurrent neural network (RNN) is conducted to address this problem, in which the hidden information will be delivered to the next step as the time passes, as shown in Fig. 3. $U$ and $V$ are the weight vectors for input and output, and $W$ is the weight for each time step. The calculative process of RNN at time step $t$ is described as following

$$H_t = f_1(a_t \cdot U_t + H_{t-1} \cdot W) \quad (13)$$

$$c_t = f_2(H_t \cdot V_t) \quad (14)$$

where $H_{t-1}$ and $c_{t-1}$ are decided at time step $t-1$. Note that, all the weights in RNN are initialized first, and then they are adjusted through back propagation to decrease the error.

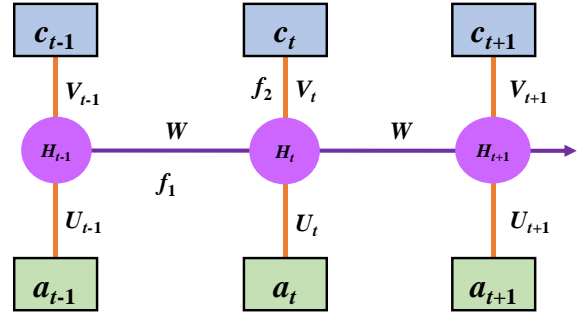

Fig. 3. The workflow diagram of RNN.

### B. Long Short Term Memory

In the backpropagation of RNN, the gradient descent algorithm is usually utilized to update the values of weight. However, the vanishing or exploding gradient problem happens frequently to stop RNN from being used in long range time sequence. As an alternative, long short term memory (LSTM) contains a memory cell to store the extra information.

The diagram of the LSTM network is depicted in Fig. 4 [16]. Six key components are included in this network, which are forget gate $F$, candidate layer $G$, input gate $I$, output gate $O$, hidden state $H$, and memory state $M$. The inputs of this network at time step $t$ are the past hidden and memory states, and the current input $X_t$. The outputs are the fresh memory and hidden states. By transiting the memory state, the LSTM network is able to remember arbitrary time intervals' series. The signs of $\sigma$ and $tanh$ denote the sigmoid and Tanh activation function, respectively. $W$ and $U$ are the weights for three gates and candidate layer with special subscripts.

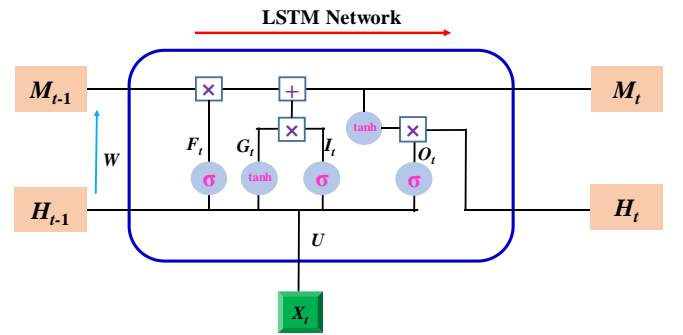

Fig. 4. The construction of the LSTM network.

According to the RNN's computational process, the expressions of LSTM can be summarized as follow

$$F_t = \sigma(X_t \cdot U_f + H_{t-1} \cdot W_f) \quad (14)$$

$$G_t = \tanh(X_t \cdot U_c + H_{t-1} \cdot W_c) \quad (15)$$

$$I_t = \sigma(X_t \cdot U_i + H_{t-1} \cdot W_i) \quad (16)$$

$$O_t = \sigma(X_t \cdot U_o + H_{t-1} \cdot W_o) \quad (17)$$

Finally, the outputs of the LSTM at time step *t* can be expressed as a function of the resulted variables

$$M_t = F_t \cdot M_{t-1} + I_t \cdot G_t \quad (18)$$

$$H_t = O_t \cdot \tanh(M_t) \quad (19)$$

After training the network with collected experienced experiment data, the learned model could provide new predictions according to new inputs. In this work, three methods for time sequence prediction are realized in Matlab 2018b. The open-sourcing NGSIM dataset is regarded as the experimental data and the details will be discussed in the next section.

*C. NGSIM Dataset*

To evaluate the performance of the proposed three approaches for time series prediction, the vehicle velocity is considered as a case study. Next Generation SIMulation (NGSIM) program was launched to collect vehicle trajectory data for research of microsimulation system [17]. The contained information is comprehensive, which consists of the speed, lane number, acceleration, vehicle type, frame numbers, position, vehicle length and so on. These driving data are recorded in four different places, which are eastbound I-80 in Emeryville, Peachtree Street in Atlanta, Lankershim Boulevard in Los Angeles and southbound US 101.

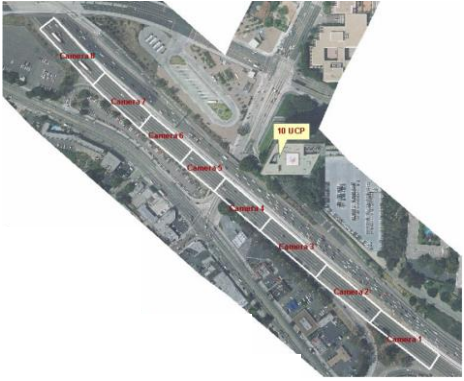

Fig. 5. The study area of the US 101 dataset in the NGSIM program.

The US 101 dataset was collected on June 15th, 2005 and three-time segments are 7:50 a.m. to 8:05 a.m., 8:05 a.m. to 8:20 a.m., and 8:20 a.m. to 8:35 a.m [18]. These periods represent the congested and uncongested driving conditions. The research area was about 640 meters and included five regular lanes, as displayed in Fig. 5. These data were collected using eight video cameras and the vehicle trajectories were transcribed by a customized software application. The recorded frequency is 10 Hz.

In this article, the speed trajectory of different vehicles is used to estimate the one-step and multi-step prediction algorithms. This information could reflect the actual traffic situations and contains random characteristics. As the NN and FC methods are both dependent on the transition probability of the driving data, they will be compared to certify the predicted performance. Furthermore, one-step and multi-step performances of the LSTM network are compared and discussed, and the root mean square error (RMSE) is used to quantify the error between the observed and predicted values

$$RMSE = \sqrt{\sum_{i=1}^{N}(x_{pre}^i - x_{ob}^i)^2 \Big/ N} \quad (20)$$

where $x_{pre}$ and $x_{ob}$ are the forecast and observed velocity trajectories, respectively.

IV. RESULTS AND COMPARATIVE ANALYSIS

This section discusses the predicted performance of the proposed three approaches. First, the importance of the historical data in NN is illuminated, which indicates the NN method needs to rely on the empirical data to achieve accurate predictions. Then, a comparative analysis between NN and FC is conducted, and their merits and demerits are described. Finally, the performance of the LSTM network is displayed by showing one-step and multi-step predicted results.

*A. NN-Based Predicted Results*

To apply the NN algorithm to forecast the speed trajectory extracted from the US 101 dataset, the transition probability is initialized as 0 before the prediction. As the time step increases, the transition probability matrix becomes more and more mature. Fig. 6 depicts three rounds prediction for the same cycle, which means NN method is used to forecast this cycle for three times. It is obvious that the performance of the second round is close to that of the third round, and they are better than the first round. In some cases, the errors are very large in the first round and they decrease in the second and third round. This owes to the update of the transition probability matrix, which indicates the probability is not stable (still the initial values) in the first round.

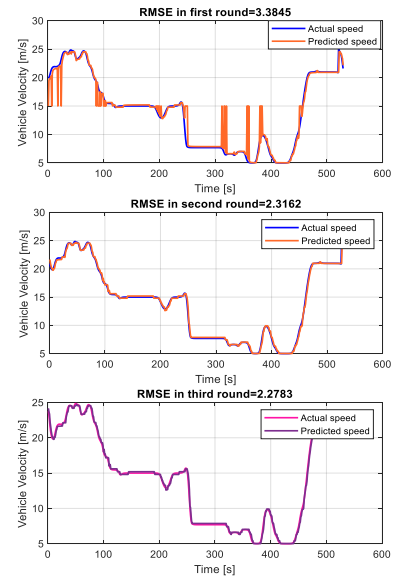

Fig. 6. Three rounds prediction using NN for one same cycle.

For multi-step prediction, Fig. 7 displays the NN-based 10-step prediction in the first and second rounds. The blue line is the observed value, and the red lines are the 10-step predicted values at each time step. By comparing the RMSE values, it can be discerned that the second round's performance is better than the first round. Hence, the NN method needs similar experienced data to complete the transition probability. After that, it can achieve good performance in one-step and multi-step predictions.

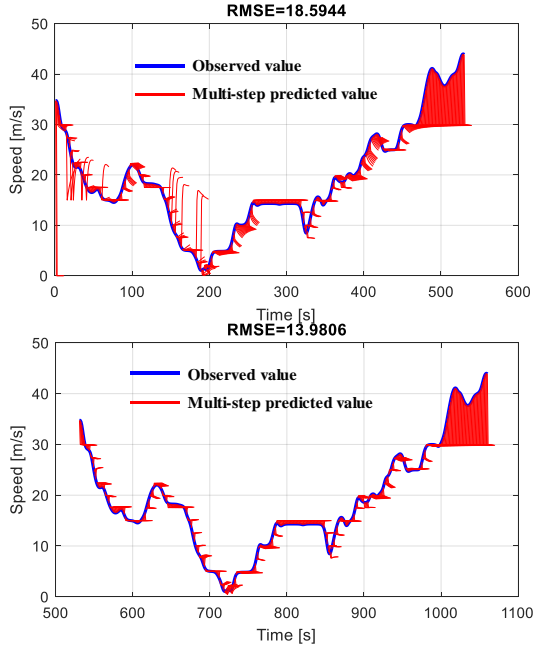

Fig. 7. NN-based 10-step prediction in the first and second round.

### B. Comparison Between NN and FC

Since the transition probability is necessary for NN and FC, they are compared in this section to declare the differences. The one-step forecast speed trajectories of these two methods are depicted in Fig. 8. Based on the curves and RMSE, it can be noted that the performance of FC is better than NN's, even in the first round. This is contributed by the division of the state space $X$ in these two methods. In NN, $X$ is divided into serval discrete states, and thus one-step speed transition only leads to one row's update in probability matrix. However, for FC, $X$ is segmented into many fuzzy sets, which results in the update of the whole probability matrix at each speed transition.

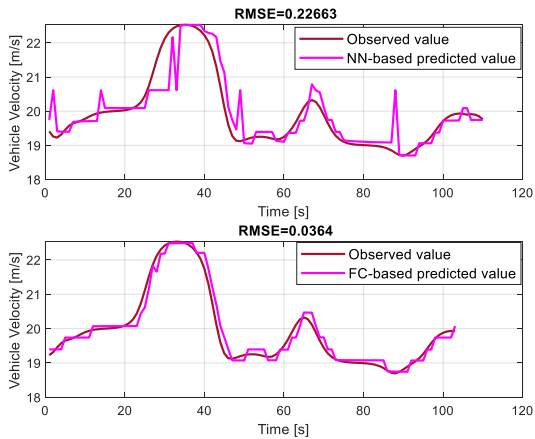

Fig. 8. Compared trajectories between NN and FC in the first round.

To intuitively exhibit the differences between these two methods, Fig. 9 shows the transition probability matrix. For NN, some high-speed transitions have not been experienced, and thus they are still initial values. Oppositely, the matrix of FC becomes stable owing to the updating rule. Besides the predicted performance, the FC approach has a drawback when compared with NN. The computation time of NN is about 2 seconds, however, it will cost 500 seconds to finish prediction in FC. Hence, FC cannot be realized in real time. To forecast some variables for autonomous vehicles, the NN method is feasible to predict the known driving situations online where the trajectory has been experienced, and the FC is appropriate to predict the unknown environments offline to achieve higher accuracy.

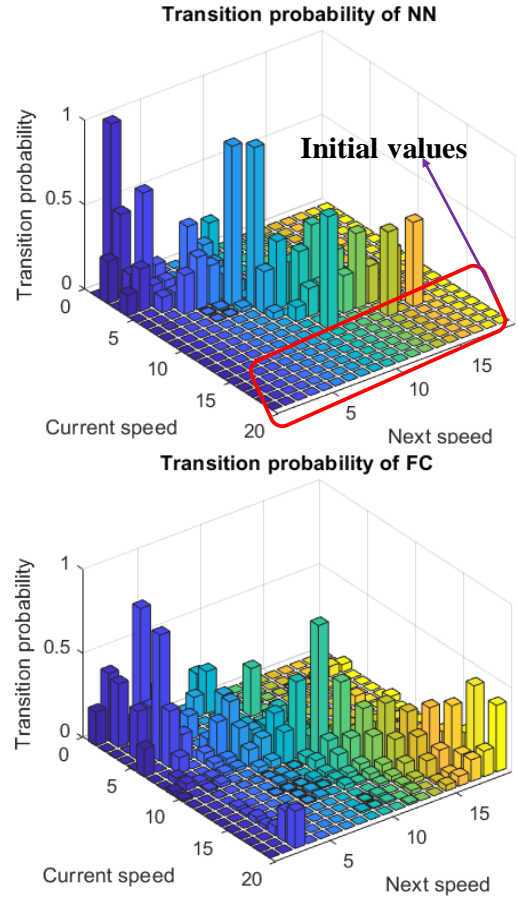

Fig. 9. Computed transition probability in two methods.

### C. Evaluation of LSTM Network

LSTM network for prediction is realized by using the *predictAndUpdateState* function in deep learning toolbox. In the operational process, the collected driving data is imported to train the network first, and then this learned model can generate new prediction by giving new inputs. The specialization of the parameters in LSTM is defined, where the training episode is 150, the number of hidden units is 100, the learning rate is 0.005 and the gradient threshold is 1. Fig. 10 describes the training process of LSTM network and it contains trajectories of RMSE and loss function. These two values drop along with the iteration number, which results in improved accuracy.

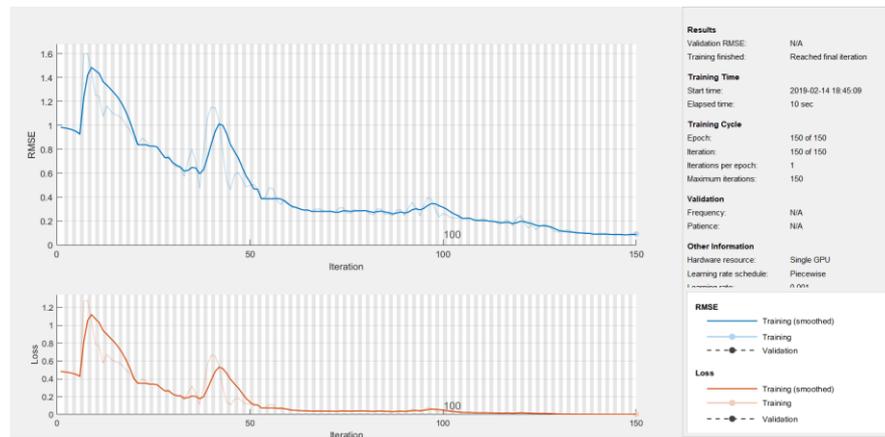

Fig. 10. Training process of the LSTM network in Matlab.

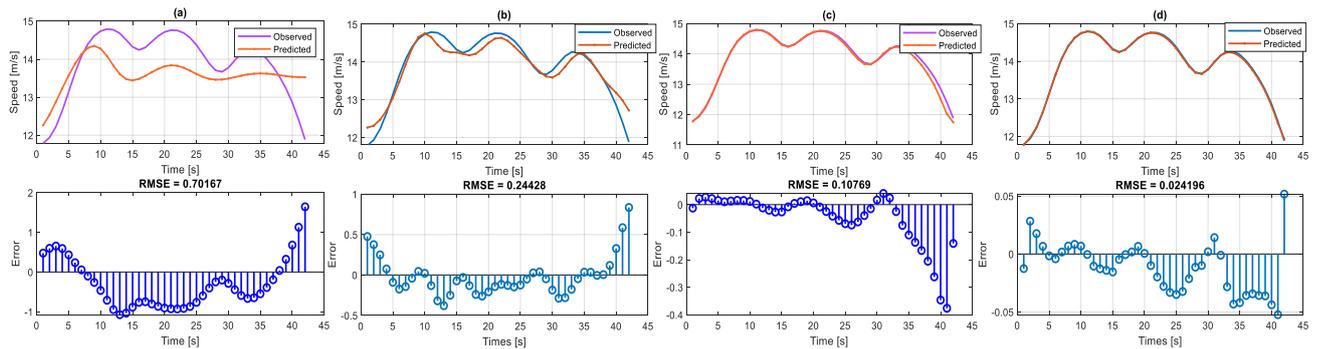

Fig. 11. The forecast results of LSTM using different training data.

To explain the significance of the training data in LSTM network, Fig. 11 shows the predicted values based on different historical data. (a) and (c) indicate the multi-step prediction (about 40 steps prediction) with different training data, and (b) and (d) are one-step prediction. It can be found that the performance of one-step is better than multi-step. It is because that the predicted errors accumulated in the multi-step prediction. Moreover, the accuracy of (c) is higher than that of (a). This can be attributed to the training data, which owes to the driving style and speed interval of the training data in (c) is close to the predicted trajectory. Therefore, when applying the LSTM network for real-time prediction, the experiment data need to be chosen properly, which means the highway driving data are better for time sequence prediction in highway scenario, and urban driving data are more suitable for urban environments.

REFERENCES

[1] F. Wang, N. Zheng, D. Cao, C. Martinez, L. Li, and T. Liu, "Parallel driving in CPSS: a unified approach for transport automation and vehicle intelligence," IEEE/CAA Journal of Automatica Sinica, vol. 4, no. 4, pp. 577-587, 2017.

[2] X. Hu, L. Chen, B. Tang, D. Cao, and H. He, "Dynamic path planning for autonomous driving on various roads with avoidance of static and moving obstacles," Mechanical Systems and Signal Processing, vol. 100, pp. 482-500, 2018.

[3] J. Wei, J. Dolan, and B. Litkouhi, "A prediction-and cost function-based algorithm for robust autonomous freeway driving," 2010 Intelligent Vehicles Symposium (IV), pp. 512-517, June 2010.

[4] S. Lefèvre, D. Vasquez, and C. Laugier, "A survey on motion prediction and risk assessment for intelligent vehicles," ROBOMECH journal, vol. 1, no. 1, pp. 1, 2014.

[5] G. Xie, H. Gao, L. Qian, B. Huang, K. Li, and J. Wang, "Vehicle trajectory prediction by integrating physics-and maneuver-based approaches using interactive multiple models," IEEE Trans. Ind. Electron., vol. 65, no. 7, pp. 5999-6008, 2018.

[6] J. Kim, and D. Kum, "Threat prediction algorithm based on local path candidates and surrounding vehicle trajectory predictions for automated driving vehicles," 2015 Intelligent Vehicles Symposium (IV), pp. 1220-1225, June 2015.

[7] J. Lee, and K. Yi, "MPC based steering control using a probabilistic prediction of surrounding vehicles for automated driving," Journal of Institute of Control, Robotics and Systems, vol. 21, no. 3, pp. 199-209, 2015.

[8] J. Wiest, M. Karg, F. Kunz, S. Reuter, U. Kreßel, and K. Dietmayer, "A probabilistic maneuver prediction framework for self-learning vehicles with application to intersections," 2015 Intelligent Vehicles Symposium (IV), pp. 349-355, June 2015.

[9] Y. Lv, Y. Duan, W. Kang, Z. Li, and F. Y. Wang, "Traffic flow prediction with big data: A deep learning approach," IEEE Trans. Intell. Transp. Syst., vol. 6, no. 2, pp. 865-873, 2015.

[10] J. Butepage, M. Black, D. Kragic, and H. Kjellstrom, "Deep representation learning for human motion prediction and classification," in Proc. IEEE Conference on Computer Vision and Pattern Recognition, pp. 6158-6166, 2017.

[11] J. Oh, X. Guo, H. Lee, R. Lewis, and S. Singh, "Action-conditional video prediction using deep networks in atari games," in Proc. Advances in neural information processing systems, pp. 2863-2871, 2015.

[12] S. Taylor, T. Kim, Y. Yue, M. Mahler, J. Krahe, er al., "A deep learning approach for generalized speech animation," ACM Transactions on Graphics (TOG), vol. 36, no. 4, 93, 2017.

[13] T. Liu, X. Hu, S. E. Li, and D. Cao, "Reinforcement learning optimized look-ahead energy management of a parallel hybrid electric vehicle," IEEE/ASME Trans. Mechatron., vol. 22, no. 4, 1497-1507, 2017.

[14] Y. LeCun, Y. Bengio, and G. Hinton, G., "Deep learning," nature, vol. 521, no. 7553, pp. 436, 2015.

[15] T. Liu, H. Yu, H. Guo, Y. Qin, and Y. Zou, "Online energy management for multimode plug-in hybrid electric vehicles," IEEE Transactions on Industrial Informatics, DOI: 10.1109/TII.2018.2880897.

[16] S. Xingjian, Z. Chen, H. Wang, D. Yeung, W. Wong, and W. Woo, "Convolutional LSTM network: A machine learning approach for precipitation nowcasting," in Proc. Advances in neural information processing systems, pp. 802-810, 2015.

[17] NGSIM Overview. Next Generation SIMulation Fact Sheet. FHWAHRT-06-135. FHWA, 2006.

[18] US Highway 101 Dataset. Next Generation SIMulation Fact Sheet. FHWAHRT-06-137. FHWA, 2006.